\documentclass{article}

\usepackage{microtype}
\usepackage{graphicx}
\usepackage{subfigure}
\usepackage{booktabs} 

\usepackage{hyperref}

\newif\iffigure

\usepackage[accepted]{icml2018}

\icmltitlerunning{Is prioritized sweeping the better episodic control?}

\usepackage{amsmath,amssymb}
\usepackage{multirow,tikz,pgfmath}

\makeatletter
\let\pgfmathrandomX=\pgfmathrandom@
\usepackage{pgfplots,pgfplotstable}
\let\pgfmathrandom@=\pgfmathrandomX
\makeatother
\newcommand{\ex}[1]{\mathbb{E}\left[#1\right]}
\pgfplotsset{compat=1.14}
\usetikzlibrary{positioning}
\usetikzlibrary{arrows}
\tikzset{
  treenode/.style = {align=center, inner sep=0pt, text centered},
  arn_n/.style = {treenode, circle, black, font=\bfseries, draw=gray!20,
    fill=gray!20, text width=1.5em, ultra thick},
}

\newcommand{\includeresults}[2]{
	\def\datafile{#1.csv}
	\def\xaxisscaling{#2}
    \def\legendname{l#1}
\begin{tikzpicture}

\pgfplotstableset{col sep=comma}
\pgfplotstableread{\datafile}\loadedtable

\pgfplotscreateplotcyclelist{mycolorlist}{%
{mgray},
{mviolet},
{myellow},
{mgreen},
{mgreen, dashed},
{mblue},
{mblue, dashed},
{morange},
{mred}
}

\pgfplotsset{compat=1.14,width=.4\textwidth, height=.3\textwidth}
\begin{axis}[
ymin=-0.02191947051959, ymax=1.04378070849877,
ytick={0.0, 0.5, 1.},
yticklabels={0.0, 0.5, 1.0},
tick align=outside,
style={font=\footnotesize},
x grid style={white},
y grid style={white},
axis line style={white},
cycle list name=mycolorlist,
xlabel={time step},
ylabel={normalized reward rate},
enlarge x limits=false,
max space between ticks=30,
scaled x ticks=real:\xaxisscaling,
legend columns=3,
legend style={font=\footnotesize},
legend entries={,,,,,,,,,,,,,,,,,,,,,,,,,,,,,,,,,,,,,,,,,,,,,Monte Carlo, n--step TD,
Q - $\lambda$, Q--learning, Q--learning with exploration, Prioritized Sweeping,
Prioritized Sweeping with
exploration, Prioritized Sweeping with reset,
Episodic Control},
legend to name=\legendname,
tick scale binop=\times,
legend style={draw=white, text=black, cells={anchor=west},
			  nodes={inner xsep=1em}}
]
\foreach \i in {1,...,5} {
	\foreach \x in {mc, nstepql, qllambda, ql, qlexplore, ps, psexplore, psreset, ec} {
		\addplot +[opacity = .2] table [x = x, y = \x_\i] {\loadedtable};
	}
}
\foreach \x in {mc, nstepql, qllambda, ql, qlexplore, ps, psexplore, psreset, ec} {
	\addplot +[line width = 1, opacity = 1.] table [x = x, y = \x] {\loadedtable};
}
\path [draw=white, fill opacity=0] (axis cs:0,-0.02191947051959)
--(axis cs:0,1.04378070849877);

\path [draw=white, fill opacity=0] (axis cs:1,-0.02191947051959)
--(axis cs:1,1.04378070849877);

\path [draw=white, fill opacity=0] (axis cs:-4.95,0)
--(axis cs:103.95,0);

\path [draw=white, fill opacity=0] (axis cs:-4.95,1)
--(axis cs:103.95,1);

\end{axis}

\end{tikzpicture}

}

\title{Is prioritized sweeping the better episodic control?}

\author{
  Johanni Brea\\
  Laboratory of Computational Neuroscience\\
  École polytechnique fédérale de Lausanne\\
  CH-1015 Lausanne \\
  \texttt{johanni.brea@epfl.ch} \\
}

\usepackage{color}
\definecolor{mblue}{HTML}{0992F2}
\definecolor{morange}{HTML}{ff7f0e}
\definecolor{mgreen}{HTML}{2ca02c}
\definecolor{mred}{HTML}{d62728}
\definecolor{mviolet}{HTML}{9467bd}
\definecolor{mbrown}{HTML}{8c564b}
\definecolor{mpink}{HTML}{e377c2}
\definecolor{mgray}{HTML}{7f7f7f}
\definecolor{myellow}{HTML}{bcbd22}
\definecolor{mcyan}{HTML}{17becf}

\definecolor{darkgreen}{rgb}{0,.5,0}
\definecolor{lightgreen}{rgb}{.5,.85,.5}

\begin{document}
\figuretrue
\twocolumn[
\icmltitle{Is prioritized sweeping the better episodic control?}




\begin{icmlauthorlist}
\icmlauthor{Johanni Brea}{epfl}
\end{icmlauthorlist}

\icmlaffiliation{epfl}{
    School of Computer and Communication Sciences and Brain Mind Institute, School of Life Sciences,
  École polytechnique fédérale de Lausanne,
  CH-1015 Lausanne}

\icmlcorrespondingauthor{Johanni Brea}{johanni.brea@epfl.ch}

\icmlkeywords{Machine Learning, ICML, Reinforcement Learning}

\vskip 0.3in
]



\printAffiliationsAndNotice{}  


\begin{abstract}
    Episodic control has been proposed as a third approach to reinforcement
    learning, besides model-free and model-based control, by analogy with the
    three types of human memory. i.e. episodic, procedural and semantic memory.
    But the theoretical properties of episodic control are not well
    investigated.  Here I show that in deterministic tree Markov decision
    processes, episodic control is equivalent to a form of prioritized sweeping
    in terms of sample efficiency as well as memory and computation demands.
    For general deterministic and stochastic environments,  prioritized sweeping
    performs better even when memory and computation demands are restricted to
    be equal to those of episodic control.  These results suggest
    generalizations of prioritized sweeping to partially observable
    environments, its combined use with function approximation and the search
    for possible implementations of prioritized sweeping in brains.
\end{abstract}

\section{Introduction}
Single experiences can drastically change subsequent decision making.
Discovering a new passage in my home town, for example, may immediately
affect my policy to navigate and shorten path lengths. A software developer may
discover a tool that increases productivity and never go back to the old
workflow. And on the level of the policy of a state, Vasco Da Gamma's discovery
of the sea route to India had an almost immediate and long-lasting effect for
the Portuguese.  

Given the importance of such single experiences, it is not surprising that
humans and some animal use an episodic(-like) memory system devoted to single
experiences \cite{Lengyel2008,Clayton2007A}. But how exactly should episodic
memory influence decision making? \citet{Lengyel2008} proposed
``episodic control'', where
\begin{quote}
	[..] each time the subject experiences a reward that is considered large
	enough (larger than expected a priori) it stores the specific sequence of
	state-action pairs leading up to this reward, and tries to follow such a
	sequence whenever it stumbles upon a state included in it. If multiple
	successful sequences are available for the same state, the one that yielded
	maximal reward is followed.
\end{quote}
This form of episodic control was found to be beneficial in the initial phase of
learning for stochastic tree Markov Decision Processes (tMDP) \cite{Lengyel2008}
and in the domain of simple video games \cite{Blundell2016}. But it is unclear, if
there are conditions under which episodic control converges to a (nearly)
optimal solution and how fast it does so.

An alternative way to link episodic memory to decision making is to compute
Q-values at decision time based on a set of retrieved episodes using a mechanism
similar to N-step backups \cite{Gershman2017}. In fact, if all episodes
starting in a given state-action pair are retrieved the result is equivalent to
TD-learning with N-step backups and decaying learning rates. More interesting is
the case where the retrieved episodes are cleverly selected, but how this
selection mechanism should work exactly, remains an open question. 

While speed of learning in the sense of sample efficiency is one important
aspect of a learning algorithm, a second criterion is computational
efficiency \cite{Lengyel2008}. Without this second desideratum, the canonical
choice would be model-based reinforcement learning, where the agent additionally
learns how its perception of the environment changes in response to actions and
how reward depends on state-action pairs.  Model-based reinforcement learning is
well known for high sample efficiency \cite{Sutton2018}. But it is often
considered to be computationally challenging. In fact, one demanding way to use
the model is to perform a forward search before every decision, similar to
chess players or board game playing algorithms \cite{Silver2017A} that plan 
the next moves. Instead of forward search at decision
time, however, one can use backward search whenever an important discovery is
made. Prioritized sweeping \cite{Moore1993,Peng1993}, implements this
backward search efficiently, in particular with small backups
\cite{Seijen2013}. 
Prioritized sweeping converges to the optimal policy for arbitrary Markov
Decision Processes (MDP) \cite{Seijen2013}. Furthermore,
for deterministic environments or at the initial phase of learning in stochastic
environments, prioritized sweeping relies in a similar way on single experiences
as episodic control. 

In this paper I address the question of whether prioritized sweeping or episodic
control are preferable in terms of sample efficiency and computational
efficiency. I introduce a variant of prioritized sweeping with model reset at
the end of each episode and show that it has the same computational complexity
as episodic control, equivalent sample efficiency in deterministic tree
Markov Decision Processes and better sample efficiency in general deterministic
MDPs or stochastic MDPs. Thus, it appears as a third and promising candidate
algorithm to link episodic memory to decision making.

\section{Review of prioritized sweeping}
Prioritized sweeping \cite{Moore1993,Peng1993,Seijen2013} is an
efficient method for doing approximate full backups in reinforcement learning. A
typical setting of reinforcement learning consists of an agent who observes
state $s_t$, performs action $a_t$ and receives reward $r_t$ at each time step
$t\in\mathbb N$. It is assumed that the state transitions and reward emissions
are governed by a stationary stochastic process with probabilities $T_{sas'} = P(s_{t+1} =
s'|s_t = s, a_t = a)$ and $R_{sa} = \ex{r_t |s_t = s, a_t = a}$. The
agent's goal is to find a policy $a = \pi^*(s)$ that maximizes 
the expected future discounted reward $V_s^\pi = \ex{\sum_{s = 0}^T \gamma^s
r_{s}\big| s_0 = s, \pi}$ for each state $s$, 
where $\gamma \in [0, 1]$ is a discount factor. With
$Q$-values $ Q_{sa}^\pi = \ex{\sum_{t=0}^\infty \gamma^t r_t\big|s_0 = s, a_0 =
a, \pi}$, we see that the optimal policy $\pi^*$ should satisfy the Bellman equations $
V_s^{\pi^*} = \max_{a} Q_{sa}^{\pi^*}$. Dropping for notational simplicity the
conditioning on the policy and using the definition of the $Q$-values, this can
be written as
\begin{equation}
    Q_{sa} = R_{sa} + \gamma \sum_{s'} T_{sas'} \max_{a'}Q_{s'a'}\, .\label{eq:bellman}
\end{equation}
From this equation, we see that a change in $Q_{s'a'}$ affects the values
$Q_{sa}$ of possible predecessor states $s$, if and only if $Q_{s'a'} =
\max_{a''} Q_{s'a''}$ before or after the change.

\begin{algorithm}
    \caption{Prioritized sweeping \cite{Seijen2013}}
    \begin{algorithmic}[1]
        \STATE For all $a, s$ initialize $Q_{sa}, V_s, U_s$, 
        all variables needed to estimate $T_{sas'}$ and an empty priority queue $PQueue$
        \FOR{each episode}
        \STATE Observe $s$
        \FOR{each step in an episode}
            \STATE Take action $a$ given by policy $\pi(Q, s)$
            \STATE Observe $r$, $s'$
            \STATE Update $Q_{sa}$ and $T_{sas''}$ (for all $s''$) given $s, a, s', r$
            \STATE $V_s \leftarrow \max_b Q_{sb}$
            \STATE Add $s$ to $PQueue$ with some priority $p$
            \STATE $s \leftarrow s'$
            \LOOP[for some steps or on background thread]
            \STATE Remove highest priority state $s^*$ from $PQueue$
            \STATE $\Delta V \leftarrow V_{s^*} - U_{s^*}$
            \STATE $U_{s^*} \leftarrow V_{s^*}$
            \FORALL{$(\bar s, \bar a)$ with $T_{\bar s\bar as^*}>0$}
            \STATE $Q_{\bar s\bar a} \leftarrow Q_{\bar s\bar a} + \gamma 
            T_{\bar s\bar as^*} \Delta V$
            \STATE $V_{\bar s} \leftarrow \max_b Q_{\bar sb}$
            \STATE Add $\bar s$ to $PQueue$ with some priority $p$
            \ENDFOR
            \ENDLOOP
        \ENDFOR
        \ENDFOR
    \end{algorithmic}\label{alg:prioritizedsweeping}
\end{algorithm}

In reinforcement learning the agent is assumed to not know the true values of
the parameters $R_{sa}$ and $T_{sas'}$. The idea of prioritized sweeping is to
maintain an estimate of $T_{sas'}$ and update the $Q$-values by interaction with
the environment and cleverly prioritized background planning.
The observation below
\autoref{eq:bellman} that limits the impact of $Q$-value changes has an important
implication in the reinforcement learning setting: if an agent, who acts in a
large, familiar environment, discovers a novel value for $R_{sa}$ or $T_{sas'}$,
this may change the $Q$-values of some predecessor states of $s$, but generally
it does not lead to a change of all $Q$-values. Hence, a model-based approach
that recomputes all $Q$-values, like value iteration, is computationally
inefficient in the reinforcement learning setting. More precisely, let us assume
\autoref{eq:bellman} holds for all state-action pairs $(s, a)$ at a given moment
in time. Next, the agent experiences in state $s$ and for action $a$ 
an unexpected immediate reward or an unexpected transition that suggest a change 
$\Delta Q_{sa} \neq 0$. If this change also alters the
value $\Delta V_s \neq 0$, all $Q$-values of predecessor states $\bar
s$ with $T_{\bar s\bar a s}>0$ should be updated to $Q_{\bar s\bar a} \leftarrow
Q_{\bar s \bar a} + \gamma T_{\bar s\bar a s} \Delta V$ (c.f.
\autoref{alg:prioritizedsweeping} line 16). This process of
updating $Q$- and $V$-values of predecessor states continues until there are
no changes in $V$-values anymore or the time budget is spent. The order in which
predecessor states are updated can be heuristically prioritized, e.g. $p = |V_s
- U_s|$ in lines 9 and 18 of \autoref{alg:prioritizedsweeping} \cite{Seijen2013}, 
such that large changes propagate faster, but also other priorization heuristics
are possible and sometimes advisable.

Empirically it is found that a small and constant number of $Q$-value updates in
the order of their priority is sufficient to learn considerably faster, i.e.
requiring fewer interactions with the environment, than model-free methods like
TD-learning or policy gradient learning \cite{Seijen2013}. This increase in
sample efficiency comes at the cost of slightly higher computation demands and a
memory complexity that is higher in general, but equal for deterministic
environments. The memory complexity of prioritized sweeping is dominated by
storing the transition table $T_{sas'}$, which, generally, has complexity
$\mathcal{O}(AS^2)$, with $A$ the number of possible actions and $S$ the size of
the state space. However, for deterministic environments, or stochastic
environments where the number of possible successor states scales as
$\mathcal{O}(1)$ (i.e. many entries of the transition table are 0), the memory
complexity of prioritized sweeping is $\mathcal{O}(AS)$, equal to the one of
model-free methods.

Prioritized sweeping is an instance of the Dyna architecture, where models, i.e.
estimates of $T_{sas'}$, are explicitly learned through interactions with the
environment and then used to update the $Q$-values by background planning, i.e.
without interaction with the environment ("trying things in your head"
\citet{Sutton1991,Sutton2018}).  Importantly, at decision time, this form of
model-based reinforcement learning has the same computational complexity as any
model-free method that relies on learned $Q$-values.

\subsection{Domain adaptation of prioritized sweeping}
The explicit dependence on the transition table
$T_{sas'}$ allows for easily incorporating prior knowledge about the domain, an
advantage that is rarely mentioned in discussions of prioritized sweeping. In
stationary and stochastic environments, the updates in line 7 of
\autoref{alg:prioritizedsweeping} depend on visit counts $N_{sa}$ of state-
action pairs $(s, a)$ and transition counts $N_{sas'}$ (such that $T_{sas'} =
N_{sas'}/N_{sa}$). $Q_{sa}$ can be efficiently updated, without the need to
recompute the sum in \autoref{eq:bellman} \cite{Seijen2013}. For
non-stationary stochastic environments it may be better to do recency-weighted
updates via leaky integration, i.e. after experiencing the transition $(s, a,
s')$ the transition table is updated according to $T_{sas''} \leftarrow (1 -
\kappa) T_{sas''}$ for all $s''\neq s'$ and $T_{sas'}\leftarrow (1 - \kappa)
T_{sas'} + \kappa$ for some $\kappa\in(0, 1)$ that could be estimated by the
agent and does not need to be the same for all $T_{sas'}$. 
For non-stationary deterministic environments, experiencing
the transition $(s, a, s')$ should result in $T_{sas''} \leftarrow 0$ for all
$s''\neq s'$ and $T_{sas'}\leftarrow 1$. And in non-stationary environments with 
mixed degrees of stochasticity, the agent could keep track of the volatility of
transitions to determine whether to treat them as intrinsically stochastic or
non-stationary deterministic.

\section{Results}
\subsection{Episodic control is equivalent to prioritized sweeping 
for deterministic tree Markov Decision Processes}
\iffigure
\begin{figure*}
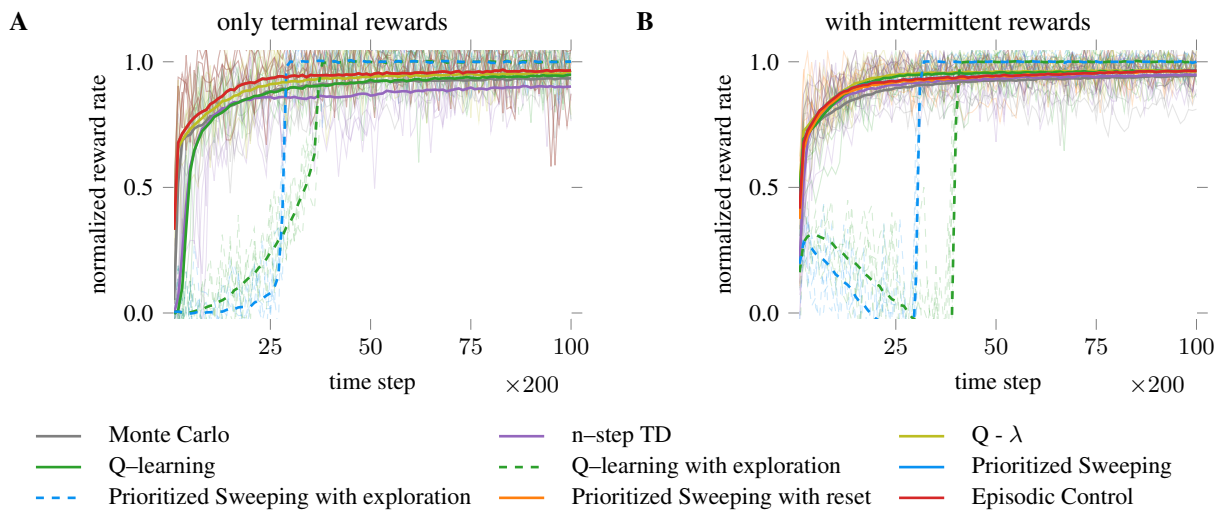

%
 \centering
%
\begin{tabular}{lclc}
		\textbf{A} & only terminal rewards & \textbf{B} &
			with intermittent rewards \\
		 & \includeresults{res1}{200} & 
		 & \includeresults{res2}{200}
\end{tabular}

\ref*{lres1}
\caption{\textbf{Deterministic tree Markov Decisions
Processes} 
\textbf{A}
The curve of the average normalized reward rate of episodic control (red) almost 
entirely covers the curves for prioritized sweeping
(blue) and prioritized sweeping with model resets (orange), indicating that they
perform equally well on deterministic tMDPs with 4 actions, depth 5 and reward
only in terminal states. The curves of N-step TD (dark blue) and  Monte Carlo
learning (gray) are very similar, as they are up to the learning rate schedule 
equivalent in
this case; being on-policy methods, they perform slightly worse. Because the
Q-learner (green) bootstraps only when re-experiencing a given state-action
pair, it learns slower, even with eligibility traces (Q-$\lambda$, yellow).
Methods with a strong exploration bias caused by large initial Q-values
("optimistic initialization", dashed curves) sacrifice initial reward for
optimal exploitation at a later stage of learning.  \textbf{B} With intermittent
rewards the differences between the algorithms almost vanish, except for the
ones with a strong exploration bias. See \autoref{sec:simulation} for
details.}\label{fig:dettmdp} 
\end{figure*}

\fi
For deterministic Markov decision processes with  $s'=T(s, a)$ and 
rewards $P(r_t|s_t = s, a_t = a) = R_{sa}$ the Bellman equations reduce to
\begin{equation}
	Q(s, a) = R_{sa}  + \gamma \max_{a'}Q\big(T(s, a),
	a')\big)\, ,\label{eq:bellmandet}
\end{equation}
where the alternative notation $Q(s,a) = Q_{sa}$ is used for readability.
In tree Markov Decision Processes (c.f. \autoref{fig:stochtmdp}A) each state
has only one predecessor state and thus
prioritized sweeping never branches during the backups. For the first episode
$\left\{s^{(1)}_1, a^{(1)}_1, r^{(1)}_1,\dots,s^{(1)}_d,
a^{(1)}_d,r^{(1)}_d\right\}$ in a deterministic tMDP of depth $d$, prioritized
sweeping may proceed as follows: no states are added to the priority queue until
the end of the episode at which point the final state $s_d^{(1)}$ is added to
the priority queue and $d-1$ backup steps (iterations of the loop
starting on line 11 in \autoref{alg:prioritizedsweeping}) are performed. After
this, the $Q$-values are given by
\begin{equation}
	Q\left(s^{(1)}_t, a^{(1)}_t\right) = G^{(1)}_t 
	\equiv\sum_{\tau=t}^d\gamma^{\tau-t}r^{(1)}_\tau\, ,
\end{equation}
if the initialization was $Q_{sa} = V_s = U_s = 0$. Note that an unconventional
but efficient use of the priority queue is made here: if states were also added
to the priority queue during the episode, the final result would be the same, if
more backup steps would be allowed (up to $\sum_{k=1}^{d-1} k = d \cdot
(d-1)/2$).
When in  episode $i$ at time step $t$ a state-action pair
$\left(s^{(i)}_t, a^{(i)}_t\right)$ is re-experienced, the backups of prioritized sweeping
result in a change of the $Q$-value that is equivalent to
\begin{equation}
	Q\left(s^{(i)}_t, a^{(i)}_t\right) \leftarrow \max \left\{Q\left(s^{(i)}_t,
			a^{(i)}_t\right), G^{(i)}_t\right\}\,
		,\label{eq:episodiccontrol}
\end{equation}
i.e. the learning rule of episodic control
used by \citet{Blundell2016}, which is a formalization of the
proposition by \citet{Lengyel2008} quoted in the introduction.

\iffigure
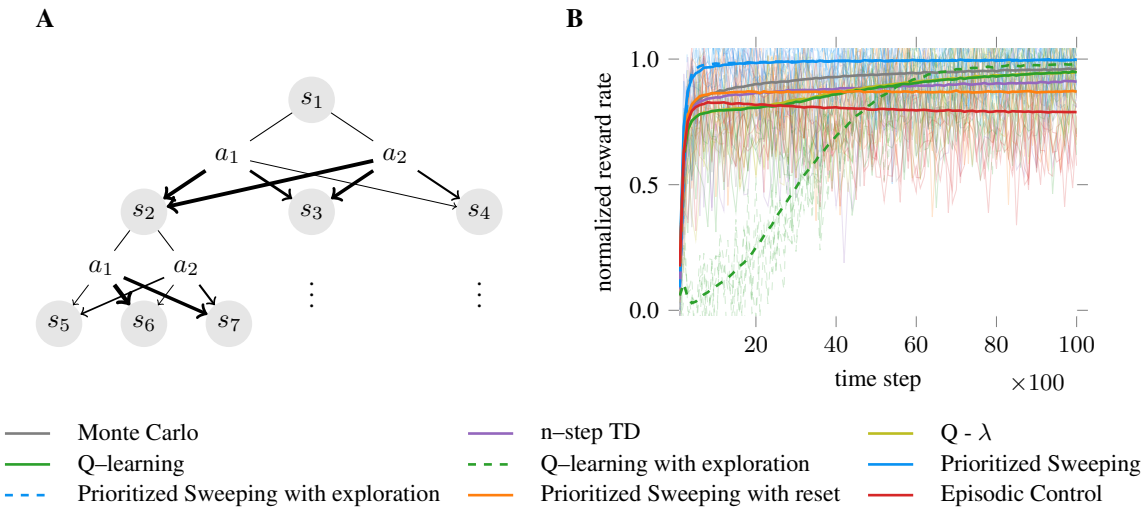
\begin{figure*}
	\centering
	\begin{tabular}{lclc}
		\textbf{A} & & \textbf{B} &  \\
			\begin{tikzpicture}[->,node distance=.3cm and .6cm]
  \node (s1) [arn_n] {$s_{1}$} ;
  \node (a1) [below left=of s1] {$a_{1}$} ; \draw [-] (s1) -- (a1); 
  \node (a2) [below right=of s1] {$a_{2}$} ; \draw [-] (s1) -- (a2);
  \node (s2) [arn_n,below left=of a1] {$s_{2}$} ;
  \node (s3) [arn_n,below right=of a1] {$s_{3}$} ;
  \node (s4) [arn_n,below right=of a2] {$s_{4}$} ;
  \pgfmathsetseed{113218}
  \foreach \i in {2,...,4}
  {	\draw [line width = 2*random()] (a1) -- (s\i); 
	\draw [line width = 2*random()] (a2) -- (s\i);}
	
	\tikzset{node distance=.3cm and .05cm};
	\node (a12) [below left=of s2] {$a_{1}$} ; \draw [-] (s2) -- (a12); 
	\node (a22) [below right=of s2] {$a_{2}$} ; \draw [-] (s2) -- (a22);
	\node (s5) [arn_n,below left=of a12] {$s_{5}$} ;
	\node (s6) [arn_n,below right=of a12] {$s_{6}$} ;
	\node (s7) [arn_n,below right=of a22] {$s_{7}$} ;
  \foreach \i in {5,...,7}
  {	\draw [line width = 2*random()] (a12) -- (s\i); 
	\draw [line width = 2*random()] (a22) -- (s\i);}
	\node (dots) [below = of s4] {$\vdots$};
	\node [below = of s3] {$\vdots$};
	\node [below = of s7] {\textcolor{white}{a}};
\end{tikzpicture}
 
			 &
				& \includeresults{res3}{100} 
				
	\end{tabular}
 	\ref*{lres3}
	\caption{\textbf{Stochastic tMDPs}. \textbf{A} Example of a stochastic
		tree MDP with depth 2, 2 actions and branching factor 3. The thickness
		of the arrow indicates the relative probability of transitioning to the
		subsequent state when choosing a given action, e.g. $T_{s_1a_2s_2} >
		T_{s_1a_2s_4}$. Rewards are in [0, 1]. \textbf{B} Learning curves for a
	stochastic tMDP with depth 4, 4 actions and branching factor 2.  See
\autoref{sec:simulation} for details.}\label{fig:stochtmdp}
\end{figure*}

\fi

To see why prioritized sweeping leads to updates equivalent to
\autoref{eq:episodiccontrol}, we note that $Q\left(s_t^{(i)}, a_t^{(i)}\right)$
will change when a novel
action $a_{t+1}^{(i)}$ is chosen\footnote{That is, 
    $a_{t+1}^{(i)} \neq a_{t+1}^{(j)}, \forall j < i$ with $s_{t+1}^{(i)} = 
s_{t+1}^{(j)}$} 
 in the subsequent state $s_{t+1}^{(i)} =
T\left(s_t^{(i)}, a_t^{(i)}\right)$ and $G_{t+1}^{(i)} >
Q\left(s_{t+1}^{(i)}, a'\right)$ for all $a'\neq a_{t+1}^{(i)}$ and thus
$\Delta V = G_{t+1}^{(i)} - G_{t+1}^{(j^*)} > 0$, where 
$j^*$ is the episode number with the previously highest return from state
$s_{t+1}^{(i)}$. Hence, by virtue of line 16 in
\autoref{alg:prioritizedsweeping}, $Q\left(s_t^{(i)}, a_t^{(i)}\right)\leftarrow
G_{t}^{(i)}$. If this leads to $\Delta V> 0$ for state $s_t^{(i)}$, prioritized
sweeping further propagates changes backwards
along the path the agent took in this episode until
$V_{s^*} = U_{s^*}$ in line 17 of
\autoref{alg:prioritizedsweeping}, after which the backup loop stops. 

In \autoref{fig:dettmdp} we see that episodic control (red curves) performs
equivalently to prioritized sweeping (solid blue curves, covered by the red
curves). Moreover, we see that episodic control and prioritized sweeping learn
faster in tMDPs than alternative methods.

For deterministic tMDPs an implementation of prioritized sweeping does not need
to maintain the transition function $T(s, a)$ beyond the end of an episode. It
has thus no influence on the performance, if the estimates of the model
parameters are reset after each episode. I call this algorithm prioritized
sweeping with model reset. In \autoref{fig:dettmdp} the performance curves of
prioritized sweeping with model reset after each episode 
(orange curves, see also \autoref{sec:simulation}) also match
the ones of episodic control (red curves).  With model reset, the memory
requirements are identical to those of episodic control. There is also no
additional cost in maintaining a priority queue, if only the last state of the
episode is added to the priority queue on line 9 in
\autoref{alg:prioritizedsweeping}, since the queue will afterwards contain
only one item at each moment in time: the predecessor of the currently backed up
state. 

We assumed above that the $Q$-values are initialized in a way such that for all
novel actions $a'$, i.e. actions that were never chosen before in state $s^*$,
$Q(s^*, a') < \min R_{sa}$. Exploration is not dramatically hampered by this
choice, if novel actions are selected whenever possible. If, on the other hand,
a maximally exploratory behavior is enforced with the initialization \linebreak
$Q(s, a) \geq d\cdot\max R_{sa}$, where $d$ is the depth of the tree, then
$\max_{a}Q\big(s, a)\big)$ may decrease over time and thus prioritized
sweeping performs also updates that are inconsistent with
\autoref{eq:episodiccontrol}. In deterministic tMDPs this leads rapidly to a
full exploration of the decision tree at the cost of low rewards in the initial
phase of learning (\autoref{fig:dettmdp} dashed blue curves). Only bootstrapping
methods, like Q-learning or prioritized sweeping, can be forced to aggressive
exploration with large initial values. Episodic control is not of this type; the
max operation in \autoref{eq:episodiccontrol} would just keep the $Q$-values
constant at the large initialization value.

\subsection{Prioritized sweeping with model reset outperforms episodic control
in general environments}
For optimal performance in stochastic (t)MDPs or in deterministic MDPs without
tree structure, prioritized sweeping needs to store the 
transition table. This means, full prioritized sweeping requires more memory and
computation than episodic control. However, an interesting competitor of
episodic control in terms of memory and computation requirements is a learner
that resets the model after each episode but uses prioritized sweeping at the
end of each episode for the
backups. If the episodes are long enough, such that the learned model for each
episode has not just tree structure, this learner may still have an advantage
over episodic control.

\iffigure
\begin{figure*}
	\centering
	\begin{tabular}{lclc}
		\textbf{A} & & \textbf{B} &  \\
				& \includegraphics[height=.33\textwidth]{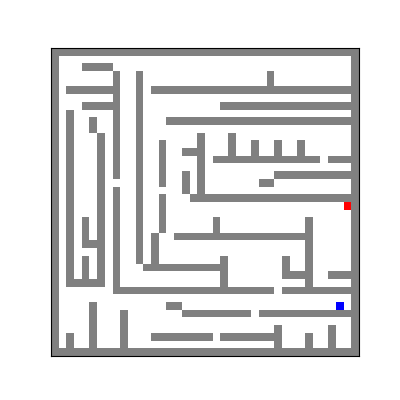} &
			 & \includeresults{res4}{10000}	
	\end{tabular}
 	\ref*{lres4}
	\caption{\textbf{Deterministic maze}. \textbf{A} Example of a deterministic 
	maze with random initial conditions (blue square) and one goal state (red
square). The gray lines indicate walls. The agent moves around by choosing
the actions up, down, left and right. If an action would move the agent into a
wall, it stays at the current position. There are usually multiple
paths that lead from start to goal. The reward is 1 for reaching the goal, 0
otherwise. \textbf{B} Prioritized sweeping with model
reset (orange curves) performs nearly optimally, while episodic control is
rapidly outperformed even by Q-learning with exploration bias. 
See \autoref{sec:simulation} for details.}\label{fig:maze}
\end{figure*}

\fi
In stochastic environments there needs to be some averaging somewhere, e.g.
explicit averaging over episodes in Monte Carlo control, choosing a small
learning rate in Q-learning or maintaining a maximum likelihood estimate of
$T_{sas'}$ in prioritized sweeping.  As already observed by Lengyel
\& Dayan \cite{Lengyel2008}, algorithms that do not properly average, like
episodic control or prioritized sweeping with model reset, may perform well at
the initial phase of learning but they lead quickly to suboptimal performance
(c.f.  \autoref{fig:stochtmdp}B, red and orange curve). While I do not see a
clear theoretical advantage of prioritized sweeping with model reset over
episodic control in such environments, it nevertheless performs slightly better in
stochastic tMDPs (see \autoref{fig:stochtmdp}B, orange above red curve), but
overall clearly worse than normal prioritized sweeping.

In deterministic environments without tree structure, states can have multiple
predecessor states. In this case, prioritized sweeping also performs backups
along \emph{virtually composed} episodes, i.e. episodes that were in their entirety not
yet experienced, but whose possibility is known to the agent from experiencing
episodes with shared states. This is similar to crossroad where the origin of
some roads is known, or the example with the uncharted passage in the
introduction. The deterministic maze environment in \autoref{fig:maze}A is of
this type: with prioritized sweeping (blue curves) the agent learns rapidly to
navigate from any starting position to the goal (red square), even if the model
is reset after each episode (orange curves). Episodic control does not
branch during the backups. While it performs well initially (red curves), it
even gets quickly outperformed by the model-free Q-learner with exploration
bonus (dashed green curves).

\subsection{Simulation details}\label{sec:simulation}
In all simulations the reward rate is measured as the mean reward obtained with
an $\epsilon$-greedy policy in adjacent windows of $T$ time steps. The
normalized reward rate is obtained by an affine transformation of the reward
rate such that 0 corresponds to the value of the uniform random policy and 1
corresponds to the policy with the optimal Q-values (obtained by policy
iteration) and the same $\epsilon$-greedy action selection. In all figures, the
thick curves are obtained by running $M$ simulations with different random seeds
on $N$ samples of the MDP in question and averaging the results. The thin curves
are 5 random samples of the $N\cdot M$ simulations. The small backups
implementation of prioritized sweeping \cite{Seijen2013} was used with 3
backups per time step. In the variant of prioritized sweeping with model reset, 
no backups were performed until the end of the episode, at
which point at most $d$ backups are performed, where $d$ is the length of the episode.
This is the same number of backups as in episodic control (c.f.
\autoref{eq:episodiccontrol}). After the end of each episode all transition
counts and observed rewards are reset to the initial value in prioritized
sweeping with reset.
The parameters of the other algorithms were chosen after a short search in the
parameter space and are probably close to but not exactly optimal for Q -
learning in stochastic tMDPs, n-step TD and Q - $\lambda$ in all environments.
See \autoref{tab:params} for the parameter choices. 
The code is available at \url{http://github.com/jbrea/episodiccontrol}.
 
\begin{table*}[t]
	\centering\footnotesize
	\begin{tabular}{r|cccccccccccc}
		\multirow{2}{*}{environment} & \multicolumn{5}{c}{shared parameters} &
		\multicolumn{2}{c}{n-step TD} & 
		\multicolumn{2}{c}{Q - $\lambda$} & Q - learning & 
		\multicolumn{2}{c}{with expl.} \\
		& $\gamma$ & $\epsilon$ & $T$ & $N$ & $M$ & $\alpha$ & $n$ & $\alpha$ &
		$\lambda$ & $\alpha$ & $Q_0^Q$ & $Q_0^{P.S}$ \\\midrule
		det tMDP \autoref{fig:dettmdp} 
        & 1.0 & 0.1 & 200 & 100 & 8 & 0.08 & 5 &
										1.0 & 0.2 & 1.0 & 5.0 & 5.0 \\
		stoch tMDP \autoref{fig:stochtmdp} 
        & 1.0 & 0.1 & 100 & 100 & 50 & 0.05
											& 5 & 0.1 & 0.2 & 0.1 & 5.0 & 5.0 \\
		maze \autoref{fig:maze} 
        & 0.99 & 0.1 & $10^4$ & 50 & 8 & 0.01 & 25 &
		0.005 & 0.2 & 1.0 & 5.0 & 0.005
	\end{tabular}\caption{\textbf{Simulation parameters.} Discount factor
	$\gamma$, probability to choose non-optimal action $\epsilon$, duration of
window to estimate reward rate $T$, number of sampled MDPs $N$, number of trials
per MDP $M$, learning rates $\alpha$, initial Q-values for Q-learning and
prioritized sweeping $Q_0^Q$ and $Q_0^{P.S}$, respectively.}\label{tab:params}
\end{table*}

\section{Discussion}
Prioritized sweeping allows to learn efficiently from a small number of
experiences. In deterministic environments the demands on memory and computation
are low and they increase gracefully with increasing
stochasticity. Episodic control is equivalent to prioritized sweeping in the
case of deterministic tree Markov Decision Processes, because in this case it is
not necessary to maintain a model beyond one episode. The observation that
episodic control performs surprisingly well in the initial phase of some Atari
games \cite{Blundell2016} may be a hint that these games are close to
deterministic tree Markov Decision Processes. 

I suggested model reset after the end of each episode. While this is a rather
ruthless form of forgetting, the simulation results show that it is still
possible to learn fairly well in the maze task. Future work will be needed to
explore more elaborate forgetting heuristics that allow to keep the memory load
low while preserving high performance.

The current formulation of prioritized sweeping requires a tabular environment,
i.e. a representation of states by integer numbers, and its superiority to
episodic control in deterministic environments becomes only evident in a
Markovian setting where states have multiple predecessors. However, tabular
environments contrast with the natural continuity of space and sensor values
encountered by biological agents and robots. Additionally, the sensory input may
contain details that are irrelevant to the task at hand, e.g. the pedestrians I
encounter every time I walk by the same place in my home town, which are
irrelevant to my task of walking from A to B.  Also, in many tasks the state is
not fully observable and state aliasing may occur, i.e.  different states have
the same observation and can only be distinguished by taking the history of the
agent into account.

This raises the question: how can we find functions that map the
sensory input to a representation useful for prioritized sweeping?
Unfortunately, prioritized sweeping does not immediately lend itself to function
approximation, in contrast to model-free learning methods like episodic control,
Q-learning or policy gradient learning \cite{Sutton2018}.  But an interesting
approach could be to learn a function that maps the high dimensional sensory
input to a discrete representation, either using unsupervised learning, e.g. a
variational autoencoder \cite{Maddison2016}, or using also reward information
to shape the map in a similar spirit as the neural episodic control model
\cite{Pritzel2017}.  

How can we generalize prioritized sweeping to partially observable
environments?  There does not seem to be an obvious answer to this question, but
memories of sequences that lead to reward together with a mechanism to attach
converging sequences could allow to go beyond updating single sequences as in
episodic control but rather make use of branching during backups as in
prioritized sweeping.  It is an open question whether this can be made to
perform better than the average over smartly selected episodes proposed by
Gershman and Daw \cite{Gershman2017}.

Prioritized sweeping relies on learning the model of the environment, i.e. the
reward table and the transition table. This may look like far from anything we
think of episodic memory. But in regions of the transition table with few
branching states, but many chains of $(s, a, s')$-triplets that are experienced
only once, backups along such chains look like recalling an episode in reversed
order. With increasing experience, it may become impossible to identify all the
episodes that contributed to the transition table. This may lead to a
gradual shift from episodic memory to semantic memory.

Is prioritized sweeping occurring in brains? The reverse replay of navigation
episodes observed in the hippocampus of rats \cite{Foster2006} can be seen as a
signature of both episodic control and prioritized sweeping. But the observation of
"never-experienced novel-path sequences" in hippocampal sharp wave ripples of
rats \cite{Gupta2010} is a feature of prioritized sweeping. If a form of
prioritized sweeping is implemented in some part of the brain of some species it
would be interesting to learn what priorization heuristics is used.

On a sufficiently abstract level, where irrelevant details of the low-level
sensory representations disappear, many tasks solved by humans seem to
be tabular and fairly deterministic and it is in this setting where prioritized
sweeping excels. This lead me to the question, if prioritized sweeping is the
better episodic control.  If the defining characteristics of episodic control is
sample efficient learning with moderate memory and computation footprint, I
would argue, the answer is yes.



\end{document}